%% file: 21-ICRA-household-marathon.tex
\documentclass[letterpaper, 10 pt, conference]{ieeeconf}  
\IEEEoverridecommandlockouts                              
\usepackage{listings}               
\usepackage[pdftex]{graphicx}       
\usepackage{xcolor}                 
\usepackage{tabularx}               
\usepackage{subcaption}
\usepackage{hyperref}               

\lstdefinestyle{Prolog} {language=Prolog,
                         lineskip=-0.5ex,
                         fontadjust=true,
                         basicstyle={\scriptsize \nopagebreak[4]},
                         commentstyle=\scriptsize,
                         moredelim=**[is][\color{easeblue}]{@}{@},
                         moredelim=**[is][\it]{~}{~}}
\lstdefinestyle{Lisp}   {language=Lisp,
                         lineskip=-0.5ex,
                         fontadjust=true,
                         basicstyle={\scriptsize \nopagebreak[4]},
                         commentstyle=\scriptsize,
                         morekeywords={achieve,def-goal,perform,def-top-level-plan,with-designators,a,an,location,object,action,at-location,with-failure-handling,
                         activity,motion,with-retry-counters,par,pursue,seq,wait-for,fail,retry},
                         moredelim=**[is][\color{easeblue}]{@}{@},
                         moredelim=**[is][\it]{~}{~}}
 \makeatletter
 \lst@CCPutMacro
     \lst@ProcessOther {"2D}{\lst@ttfamily{-{}}{-}}
     \@empty\z@\@empty
 \makeatother

\definecolor{ease_darkblue}{HTML}{144F78}
\colorlet{easeblue}{ease_darkblue}

\title{\LARGE \bf 
The Robot Household Marathon Experiment
}

\author{Gayane Kazhoyan, Simon Stelter, Franklin Kenghagho Kenfack, Sebastian Koralewski and Michael Beetz$^{*}$\\
        {\{kazhoyan, stelter, fkenghag, seba, mbeetz\}@uni-bremen.de} 
        \thanks{$^{*}$The authors are with the Institute for Artificial Intelligence, University of Bremen, Germany.}}

\begin{document}
This work has been submitted to the IEEE for possible publication. Copyright may be transferred without notice, after which this version may no longer be accessible.
\newpage
\maketitle
\thispagestyle{empty}
\pagestyle{empty}


\begin{abstract}

In this paper, we present an experiment, designed to investigate and evaluate the scalability and the robustness aspects of mobile manipulation. The experiment involves performing variations of mobile pick and place actions and opening/closing environment containers in a human household. The robot is expected to act completely autonomously for extended periods of time. We discuss the scientific challenges raised by the experiment as well as present our robotic system that can address these challenges and successfully perform all the tasks of the experiment. We present empirical results and the lessons learned as well as discuss where we hit limitations.


\end{abstract}


\section{Introduction}
\label{sec:intro}

Consider a robot that should carry out everyday chores in a household, such as setting a table or tidying up.
In the state of the art robotics, there has been a number of research endeavors towards implementing integrated robotic systems capable of executing mobile manipulation tasks under strict assumptions (see Section \ref{sec:soa}).
However, in order to be able to act competently in an arbitrary human user's household with its various objects,
the software of the robot has to be able to scale towards the unrestricted open world.

\begin{figure}[htb]
	\centering
	\includegraphics[height=0.31\columnwidth]{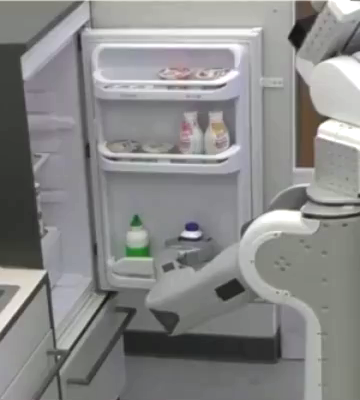}
	\includegraphics[height=0.31\columnwidth]{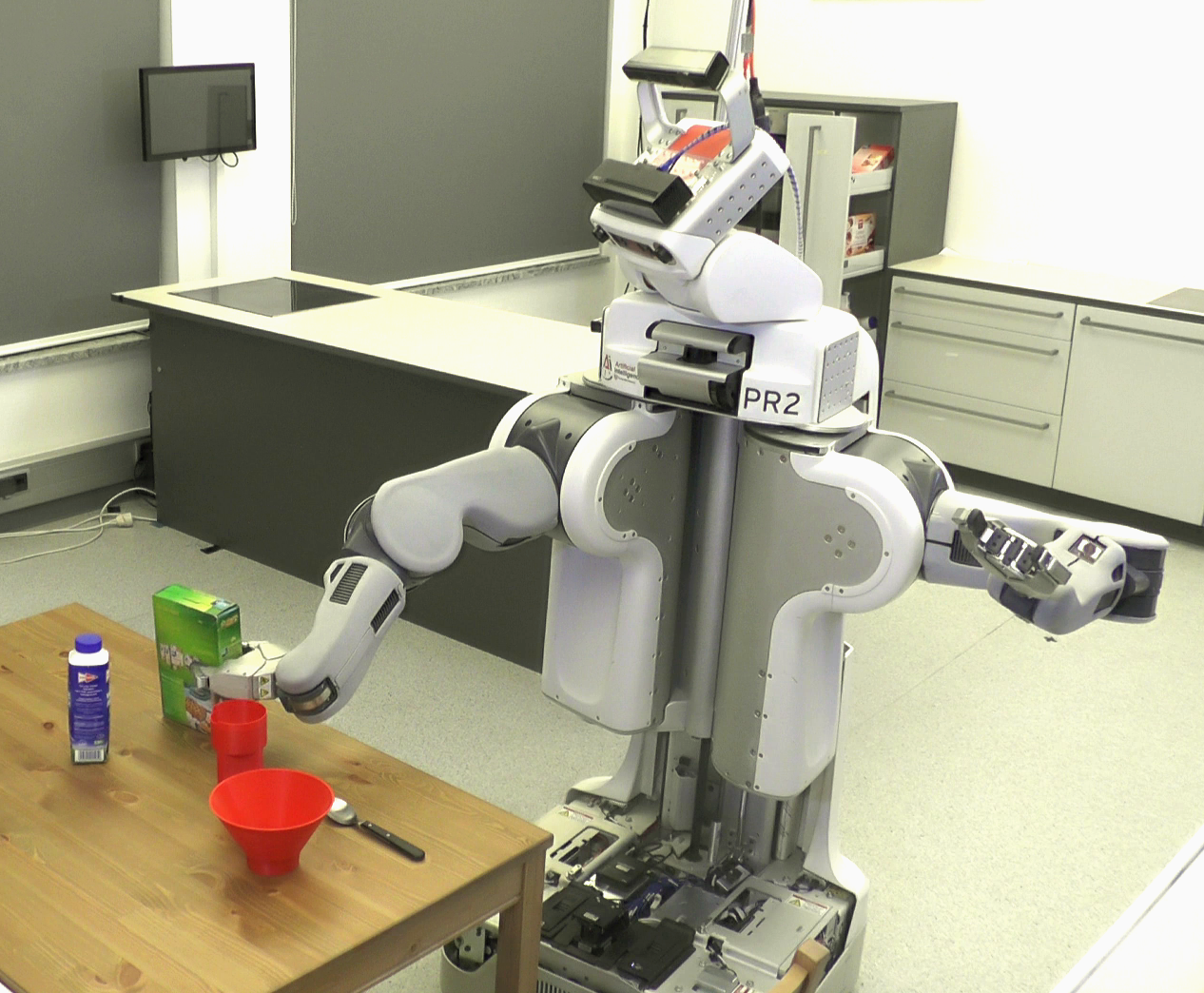}
	\includegraphics[height=0.31\columnwidth]{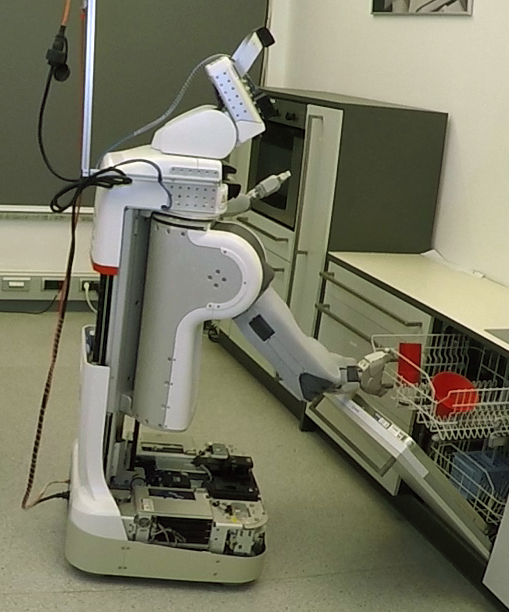}
	\caption{PR2 setting a table and cleaning it up: picking up the milk box, placing the cereal box, closing the dishwasher.}
	\label{fig:intro:cover}
\end{figure}

To investigate the scalability and the robustness aspects of mobile manipulation systems,
we designed an experiment that involves performing variations of mobile pick and place actions and opening/closing environment containers in a human household (see Figure \ref{fig:intro:cover}). The robot is required to act completely autonomously for extended periods of time. 
We call this experiment the robot household marathon\footnote{\scriptsize \url{https://www.youtube.com/watch?v=pv_n9FQRoZQ}}.

The experiment consists of two main household tasks. 
The first task is to set a table for breakfast for one person. This involves five objects -- a bowl, a spoon, a cup, a milk box and a box of cereal -- which have to be brought to the dining table. The second task is to clean the table, such that washable objects are placed into the dishwasher, the cereal box is placed back into its storage location, and the milk box is thrown away into the trash, as it is assumed to be empty after the meal. The objects vary in shapes and appearances and the manipulated containers have different articulation models. The source and destination locations of the objects are varied as well. The robot is provided with symbolic descriptions of object locations in containers and on surfaces but the exact poses change between the experiment runs. The robot freely chooses the arm to use for grasping, the grasp pose, the corresponding base pose, etc.

There are three main research challenges that have to be addressed to master these actions.
First, the robot needs to be able to execute the mobile pick and place actions and the opening/closing of doors and drawers in a wide range of variations. This means being able to get any object from any location in the environment, including locations inside containers, and place it at any other plausible location. Accomplishing such highly underdetermined tasks requires to autonomously infer the appropriate behavior in the given context.
Each component of the system has to be able to scale towards these variations, including robot's perception, motion planning, grasp planning, plan executive, etc.
The second challenge is dealing with the non-determinism and the partial observability of the world: acting in the real world very often leads to failures and the robot needs to be able to detect them and react accordingly. This requires, among other things, to keep the robot's beliefs about the world in a consistent state, as close to the real world as possible.
Finally, the experiment requires an interplay of different branches of robotics and creates system interaction challenges, which are often out of scope of specialized roboticists' research.

In this paper, we present a system that can successfully execute the household marathon experiment.
It addresses the above-mentioned challenges by following a few principles:
\begin{itemize}
	\item The parameters of each robot action are inferred by querying \textit{reasoning engines} through a well-defined API.
	\item One of our reasoning engines
	is based on heuristics that are made 
	general and scalable by sacrificing optimality.
	\item To eventually reach human-level skill, the system is equipped with an experience-based learning framework.
	The framework provides a reasoning engine for inferring action parameters based on a learned model.
	\item The actions and their goals are described with first-order logic, which allows reasoning about action execution.
	\item To ensure fully autonomous behavior, the robot action plans are equipped with failure handling constructs.
	The goal monitoring processes run concurrently with action execution, thus, the failures are detected immediately.
	\item To ensure the consistency of the internal world state representation, we use a physics engine.
	\item Our perception works under very few assumptions and is able to perceive objects that are oriented arbitrarily in space and/or have no distinct supporting plane. 
	\item We use a full-body motion planner based on a constraint and optimization approach. It allows to specify concurrent goals, e.g., ``move to a pose, while looking at your hand as well as avoiding collisions and joint limits''.
\end{itemize}


The contributions of this paper to the state of the art and our approach to addressing the challenges raised by the experiment and the implemented system.
In the remainder of this paper, we review related work, explain the integral components of our system and their interaction,
and report on the experimental results and the lessons learned.


\section{Related Work}
\label{sec:soa}

Below we outline related work in the scope of integrated autonomous robotic systems for mobile manipulation.

One of the earlier examples of a mobile manipulation robot performing tasks in the household was Okada's system on the HRP2 \cite{dishes}, which could execute tasks such as grasping a cattle or carrying a tray with objects. A later iteration of the work \cite{jsk-tidying-room} presented an assistive household robot that could load a washing machine and sweep the floor.

One of the first examples of ROS-based mobile manipulators is the home robotic butler HERB \cite{srinivasa2010herb}, which could do localization, vision and opening/closing of doors. A large amount of software systems for mobile manipulation have been developed during the PR2 beta program: \cite{pr2beer} shows a fully integrated pick and place system with vision, navigation, motion planning and grasping components;
\cite{pr2bakebot} describes the PR2 cookie baking experiment;
and our previous work \cite{pancakes} had two robots making pancakes, where the PR2 did mobile pick and place, and the TUM-Rosie robot did pouring of the batter and the flipping of the pancake.  

Outside of the household domain, systems have also been developed for performing maintenance tasks in industrial environments, e.g., the collaborative ARMAR-6 robot \cite{armar}.

Progress in mobile manipulation has also been showcased in robotics competitions.
%
One of the interesting integrated systems shown at the Robocup@Home competitions is the cognitive service robot Cosero \cite{behnkerobocup}, which in addition to perception, motion planning and navigation modules, also contains human-robot interaction functionality.
%
%
In \cite{fetch-challenge} an experiment conducted in the scope of the Fetch challenge is shown, where the goal is to assemble a mechanical kit.


The focus of this paper is, first of all, on a system that can do \textit{both navigation and manipulation}. Thus, we do not mention work that considers only the manipulation or only the mobile aspect. Second, we are looking at experiments with complex scenarios, where the robot autonomously operates for extended periods of time, which requires considerable \textit{robustness}. Finally, instead of implementing specialized solutions for each action variation, we aimed at one \textit{scalable} solution that can cover a multitude of variations. We consider our system to be different from related work and our previous work \cite{pancakes} in our approach to robustness and scalability. 


\section{System Architecture}
\label{sec:arch}

\begin{figure}[htb]
	\centering
	\includegraphics[width=\columnwidth]{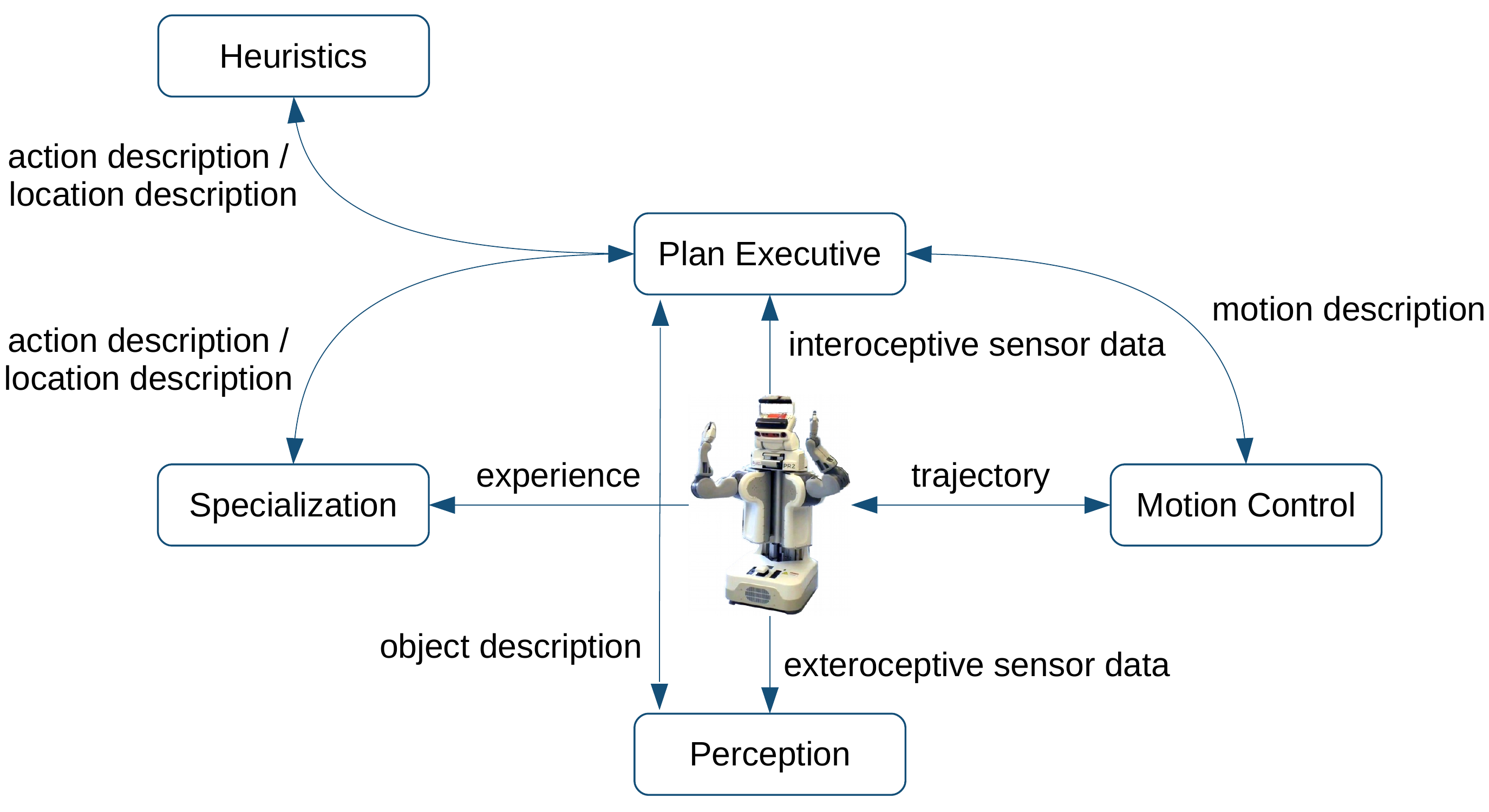}
	\caption{Architecture diagram of our robotic system.}
	\label{fig:arch:arch}
\end{figure}

Our robotic system has five main modules (see Figure \ref{fig:arch:arch}).

The Plan Executive decides on the goals of the robot and the sequence of actions to perform to achieve these goals. 
Additionally, it uses data from the optical encoders in the joints of the robot to monitor action execution. It ensures that everything goes as planned and reacts to failures. 

To infer good action parameter values that are expected to lead to successful action execution, the Plan Executive can use the Heuristics module. The input of the module is an action description with missing parameters, and the output is an augmented description with the corresponding parameter values filled in. Additionally, a symbolic location description can be sent to the Heuristics module, to be grounded to a specific pose in the robot's environment.

The Experience-based Specialization system collects experiences of the robot during execution. Simultaneously, in a continuous loop, it specializes the action parameters towards the given robot body and the given environment using reinforcement learning approaches.

The Plan Executive can use either the Heuristics module for inferring parameters, or the Specialization-based module, as they have the same API. At the beginning, the Heuristics are used for all parameter inference tasks; the robot performs the experiment and the execution data is logged. Later, when the amount of collected experience becomes sufficient for training models, for certain parameters the Heuristics module is completely replaced by the Specialization module, and for the other parameters the results from both are combined.

The Perception system receives symbolic descriptions of objects from the Plan Executive and fits them to the objects that the robot currently sees in its given environment, as registered in the RGBD data stream from the camera mounted on the robot's head. The output of the module is an augmented object description, where additional properties of the object are added, such as its exact pose, color, etc.

The Motion Planning framework receives symbolic descriptions of motions from the Plan Executive and translates them into trajectories executable on the given robot in the given environment. It then sends the trajectories to the robot's low-level controllers. 



\section{Plan Executive}
\label{sec:cram}


This section explains our Plan Executive module, called CRAM\footnote{\scriptsize \url{http://cram-system.org} \hspace{1em} \url{https://github.com/cram2/cram}}, which is responsible for the following tasks:
\begin{enumerate}
	\item define which actions to execute to achieve the goal;
  	\item infer which parameters to use for each action;
   	\item monitor task execution and react to failures.
\end{enumerate}

\paragraph*{Task 1}
To decide which actions to execute and in which order in the given situation, 
we use the concept of \textit{generalized plans}.
We call a robot action plan \textit{generalized} if the same plan can accomplish a large amount of variations of one action category.
For example, for the action category of mobile pick and place this means being able to fetch any object from any location in any environment 
and place it at any plausible destination location, including inside containers.
It should be possible to achieve all the variations with the same generalized plan.
Realistically, it is not feasible to cover every possible situation, so we aim at scalable plans that require the least amount of reprogramming effort when changing the domain.
%
%
The sequence of actions is engineered based on an accurate model of mobile pick and place. 
%


To achieve scalability, the generalized plans are written with underspecified symbolic action descriptions.
For example, in the code listing below, we command the robot to \textit{perform} an underspecified action of type \textit{picking-up} on some previously perceived object of type \textit{cup}, whereby the goal of the action is to have the cup object in the hand:

\begin{lstlisting}[style=Lisp]
(perform 
  (an action
      (type picking-up)
      (object (an object (type cup) (pose ...) ...))
      (goal (object-in-hand ~#object~))))
\end{lstlisting}

The \textit{perform} operator, which is defined in our domain-specific robot programming language library CRAM-PL \cite{cram}, gets the action description data structure as its single input argument and executes the action on the robot.

The \textit{(object-in-hand ...)} clause is written in first-order logic, which allows the robot to check if the clause holds w.r.t.\ the current world state representation. If that is the case before performing the action, executing it is omitted, as its goal has already been achieved. 

\paragraph*{Task 2} Most of the parameters of the action in the code listing above are omitted from the description,
which makes it very general.
During execution, these missing parameters are inferred at runtime by querying the reasoning engines, taking into account the given situation.
The resulting action description with inferred parameters is shown below:

\begin{lstlisting}[style=Lisp]
(an action
    (type picking-up)
    (object (an object (type cup) (pose ...) ...))
    (goal (object-in-hand ~#object~))
    (arm ~left~)
    (grasp ~identifier-of-one-of-the-grasps-for-cup-objects~)
    (gripper-opening ~finger-opening-distance-for-grasping~)
    (trajectory ~picking-up-trajectory-via-points~)
    (grasping-force ~30Nm~))
\end{lstlisting}


In the action parameter inference step, the Plan Executive queries the Heuristics module or the Specialization module, asking questions
such as "Where can I find the object of the given type in my environment?", "Where should I stand to be able to perceive this object?" etc.
%
%
%
To make sure that the returned answer is good, the robot performs the action with the inferred parameters in a fast simulator \cite{projection}, prior to real-world execution. 
If the action fails, another suitable parameter value is inferred and the action is simulated again.

\paragraph*{Task 3} Our Plan Executive uses the reactive planning approach, such that the plans perform the actions and monitor their execution concurrently. For example, if an object slips from the gripper during the carrying action, the robot immediately stops executing the trajectory and starts the failure handling strategy. For specifying the concurrent-reactive behavior, we use the operators from CRAM-PL \cite{cram}, which take care of the multithreading and synchronization under the hood. 
Below is a simplified version of the code, which implements reacting to object slipping from the gripper:

\begin{lstlisting}[style=Lisp]
(pursue (perform (an action (type carrying) (object ...)))
        (seq (wait-for ~fingers-closed-completely-event~)
             (fail ~object-slipped-error~)))
\end{lstlisting}

The \textit{pursue} operator executes its child clauses in parallel threads, until either of them finishes, at which point the other threads are terminated; \textit{seq} simply executes its children in a sequence; \textit{wait-for} blocks the thread until an event happens. 

CRAM-PL also has operators for writing failure handling strategies. The most frequently used one is to simply retry performing the action with another set of parameters. The failure handling strategies are engineered according to our general model of mobile pick and place. 

\section{Experience-based Specialization}

\input{learning.tex}
\section{Perception}
\label{sec:rs}

To perceive an object in the environment, the Plan Executive sends a vague object description to the Perception system, e.g., the description of a red cup-shaped object is:

\begin{lstlisting}[style=Lisp]
(an object (type cup) (color red))
\end{lstlisting}

\input{perception.tex}

To ensure the physical plausibility of the estimated pose, a CAD model of the object is spawned in the simulator containing the current world state, which is followed by simulating the world for $3s$. Thus, if the object was spawned inside another object, e.g., the fridge door, the pose gets corrected. If the corrected pose differs significantly from the initial pose, e.g., if the object falls to the ground when simulated, the initial pose is corrected iteratively with small offsets until a stable pose is found for the object.



\section{Motion Planning}
\label{sec:giskard}

\input{motion_planning.tex}

\section{Experimental Analysis and Discussion}
\label{sec:eval}

A full table setting and cleaning experiment takes about $90 min$.
For experimental analysis, we performed the full scenario five times and recorded robot's performance.

\begin{table}[tb]
\setlength{\tabcolsep}{0.4em}
\renewcommand{\arraystretch}{1.2}
\begin{tabularx}{\columnwidth}{|X|p{3em}|p{3em}|p{3em}|p{3em}|p{3em}|p{3em}|p{3em}|}
\hline 
\textbf{Object} & \textbf{Unre-cover.\ fail.} & \textbf{Perc.\ fail.} & \textbf{Grasp.\ fail.} & \textbf{Manip.\ fail.} &
\textbf{Env.\ manip.\ fail} & \textbf{Nav.\ fail.} & \textbf{Dura-tion ($sec$)} \\\hline
Bowl            & 0                           & 2/5                   & 0                      & 29/5                   &
0                    & 0                           & 515                        \\\hline 

Spoon           & 0                           & 5/5                   & 3/5                    & 13/5                   &
0                    & 0                           & 522               \\\hline 

Cup             & 0                           & 0                     & 2/5                    & 23/5                   &
0                    & 0                           & 472               \\\hline

Milk            & 0                           & 3/5                   & 0                      & 29/5                   &
13/5                 & 0                           & 687               \\\hline

Cereal          & 0                           & 0                     & 0                      & 10/5                   &
5/5                  & 3/5                         & 530               \\\hline

Sum             & 0                           & 10/5                  & 5/5                    & 104/5                  &
18/5                 & 3/5                         & 2726              \\\hline 
\end{tabularx}
\caption{Results of table setting, averaged over 5 runs.}
\label{tbl:setting}
	\vspace{-0.6em}
\end{table}

In Table \ref{tbl:setting} you can see the results of the table setting part of the experiment, averaged over the 5 runs.
No unrecoverable failures happened during table setting, so all the runs were successful.
The most frequent failures happened during manipulation, which had three reasons: (1) the inverse kinematics solver deemed the goal unreachable, (2) the motion planner could not generate a trajectory due to narrow space, e.g., inside a drawer and (3) the perception system gave an incorrect pose estimate, which in its turn resulted in (1) or (2).
The navigation only failed for the cereal box, as its storage is located the furthest away from the dining table, with another table standing in between, making it hard for our motion planner to find a suitable trajectory.
Perception failures that the robot registered happened the most with the spoon object, as it is flat and small. For the same reason, misgrasps happened the most with this object as well.
Environment manipulation only failed in the case of the milk and the cereal, as pulling on the handle of the refrigerator and the heavy vertical drawer required a lot of force and often resulted in the handle slipping from the gripper.
All of these failures were successfully handled by retrying the action with another set of values for its parameters.

The whole table setting part of the scenario took on average $2726s \approx 45.43min$.
Bringing the milk to the dining table was the action with the longest average duration, it also had the most failures of all types.
The quickest mobile pick and place action of one object was transporting the bowl in the run that had $0$ failures, which took $339s = 5.65min$.
Perception actions took anywhere between $2-74s$ due to the use of the ICP algorithm.
Navigation from the storage locations of objects to the dining table took on average $1min$, including the planning time.
The same amount was required to close the door/drawer of the storage.
Grasping/placing objects, as well as opening/closing doors/drawers took anywhere between $24-49s$. 

\begin{table}[htb]
\setlength{\tabcolsep}{0.4em}
\renewcommand{\arraystretch}{1.2}
\begin{tabularx}{\columnwidth}{|X|p{3em}|p{3em}|p{3em}|p{3em}|p{3em}|p{3em}|p{3em}|}
\hline 
\textbf{Object} & \textbf{Unre-cover.\ fail.} & \textbf{Perc.\ fail.} & \textbf{Grasp.\ fail.} & \textbf{Manip.\ fail.} &
\textbf{Env.\ manip.\ fail} & \textbf{Nav.\ fail.} & \textbf{Dura-tion ($sec$)} \\\hline
Cereal          & 0                           & 1/5                   & 7/5                    & 14/5                   &
4/5                         & 3/5                  & 688                        \\\hline 

Milk            & 1/5                         & 4/5                   & 3/5                    & 0                      &
14/5                        & 5/5                  & 604               \\\hline 

Spoon           & 1/5                         & 3/5                   & 2/5                    & 2/5                    &
16/5                        & 3/5                  & 656               \\\hline

Cup             & 0                           & 0                     & 4/5                    & 14/5                   &
0                           & 1/5                  & 328               \\\hline

Bowl            & 0/5                         & 0                     & 1/5                    & 19/5                   &
9/5                         & 0                    & 640               \\\hline

Sum             & 2/5                         & 8/5                   & 17/5                   & 49/5                   &
43/5                        & 12/5                 & 2916              \\\hline 
\end{tabularx}
\caption{Results of table cleaning, averaged over 5 runs.}
\label{tbl:cleaning}
\end{table}

Table \ref{tbl:cleaning} shows the averaged results of the table cleaning part of the experiment.
Cleaning the table was more challenging than table setting, due to the use of the dishwasher and the difficulty of sideways grasping objects located far away from the edge of the table.
\begin{figure}[htb]
	\centering
	\includegraphics[width=0.32\columnwidth,height=0.32\columnwidth]{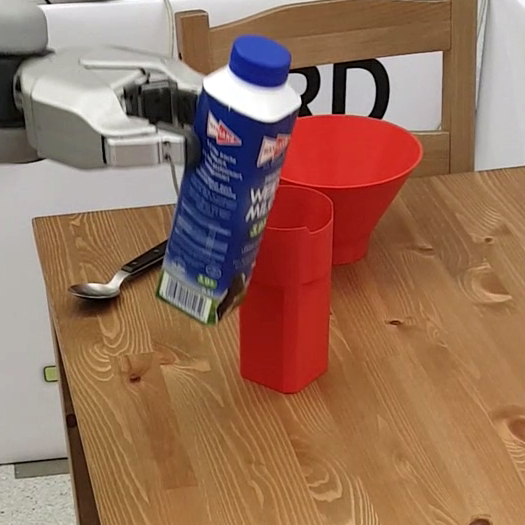}
	\includegraphics[width=0.32\columnwidth,height=0.32\columnwidth]{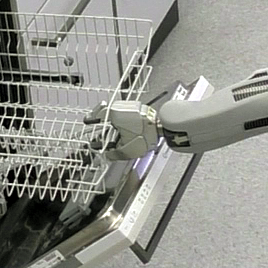}
	\includegraphics[width=0.32\columnwidth,height=0.32\columnwidth]{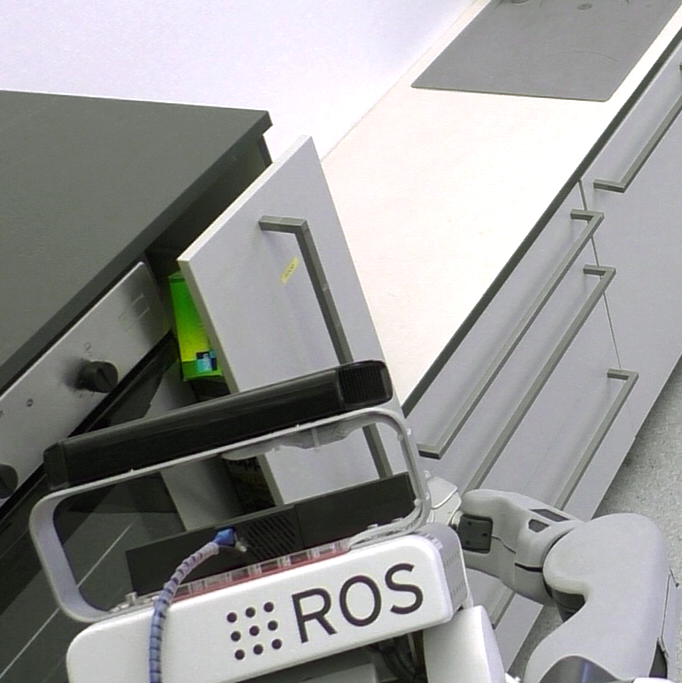}
	\caption{Some of the failures that happened during table cleaning: (left) unstable milk box grasp, which resulted in the object slipping, (middle) dishwasher grid stuck in a finger cavity, (right) cereal box squashed by the closing drawer.}
	\label{fig:eval:fails}
\end{figure}
In two out of the five runs we encountered an unrecoverable failure.
In one of the runs, due to the instability of the grasping trajectory and the robot not tracking it perfectly, the fingers of the robot ended up pushing the milk away during grasping, which resulted in a very unstable grasp (see Figure \ref{fig:eval:fails} (left)). As a result, the box fell to the ground in the carrying phase.
Although during the table setting the robot was able to pick up a toppled over cup and successfully bring it to the table, picking up the milk box from the ground was impossible for the PR2.
The other unrecoverable failure was the dishwasher grid getting stuck in PR2's finger (see Figure \ref{fig:eval:fails} (middle)). 
Another major failure happened when placing the cereal box into its vertical drawer, which was difficult because the robot had to reach very high and approach its joint limits. When the gripper opened, the box fell on a side in the shelf,
which resulted in it being crashed when the drawer was closed (see Figure \ref{fig:eval:fails} (right)). The robot still considered this run as successful, as no unrecoverable failures were registered.


While working with the robot, we observed that localization errors and imprecise trajectory tracking made the robot clumsy, such that it would knock over objects and occasionally slightly hit open drawers and doors. We do not have a possibility to register accidental changes to the state of the doors and drawers but due to failure handling and consistency checking of the world state, this did not cause major problems. Localization error was further worsened when opening large drawers that are on the level of the robot's laser scanner. Using full-body motions results in error accumulation from the additional degrees of freedom but the gains in reachability greatly outweigh that. Our visual perception works on demand and does not do tracking but monitoring the joint states of the robot compensated for that.


\section{Conclusion}
\label{sec:conclusion}

In this paper, we presented a mobile manipulation experiment that involves performing variations of mobile pick and place tasks in a household.
The experiment requires the robot to act completely autonomously in the real world for extended periods of time.
We presented a system that allows the robot to successfully execute this challenging experiment multiple times, autonomously recovering from the encountered failures.


\section*{ACKNOWLEDGEMENTS}
\begin{small}
\noindent
This work was supported by the DFG CRC \emph{Everyday Activity Science and Engineering (EASE)} (CRC \#1320), the DFG Project \emph{PIPE} (project \#322037152), the EU H2020 Project \emph{REFILLS} (project \#731590) and BMWi project \emph{Knowledge4Retail}. This experiment would not be possible without the help of Ferenc B\'alint-Bencz\'edi, Jeroen Sch\"afer, Alexis Maldonado, Patrick Mania and others.
\end{small}


\bibliography{bibliography}
\bibliographystyle{IEEEtran}

\end{document}

%% file: learning.tex
\label{sec:ralf}

To find suitable action parameters, our Plan Executive applies brute-force search over the parameter search space. A parameter is considered suitable if it results in successful execution of the action, first, in simulation and then also on the real robot. 
To keep the search feasible, we rely on domain discretization, heuristics and learned models.

\begin{figure}[htb]
	\centering
	\includegraphics[width=0.32\columnwidth,height=0.32\columnwidth]{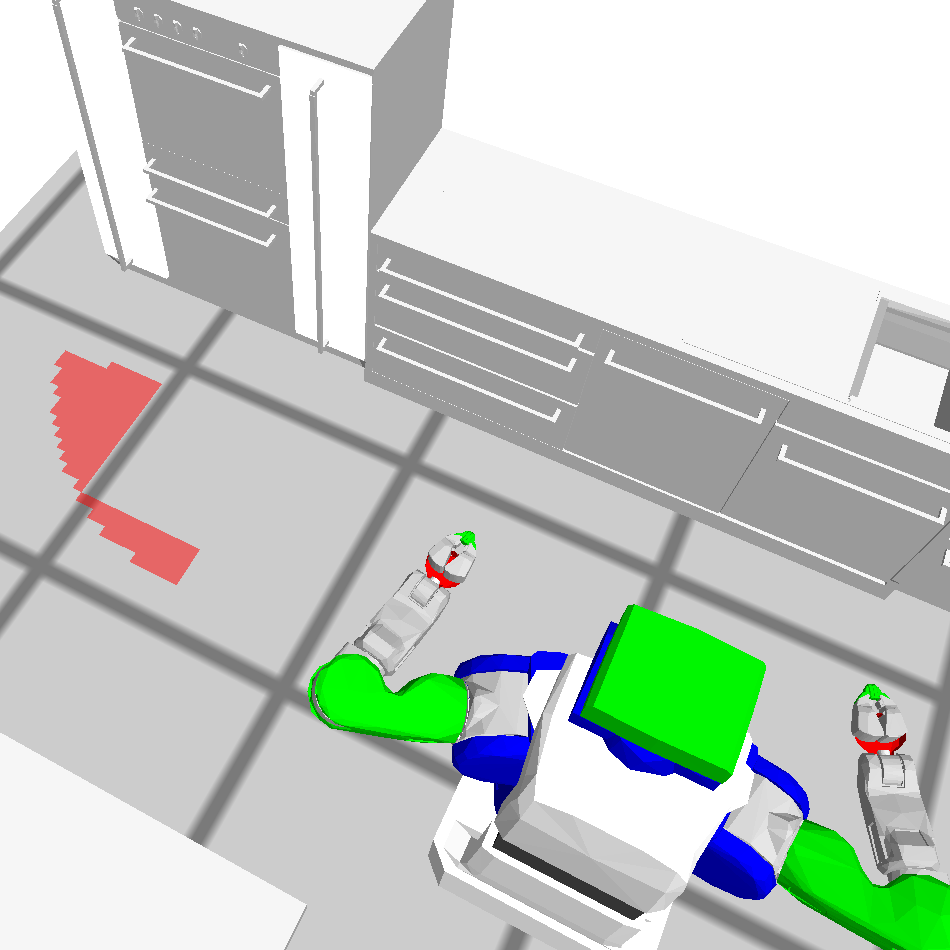}
	\includegraphics[width=0.32\columnwidth,height=0.32\columnwidth]{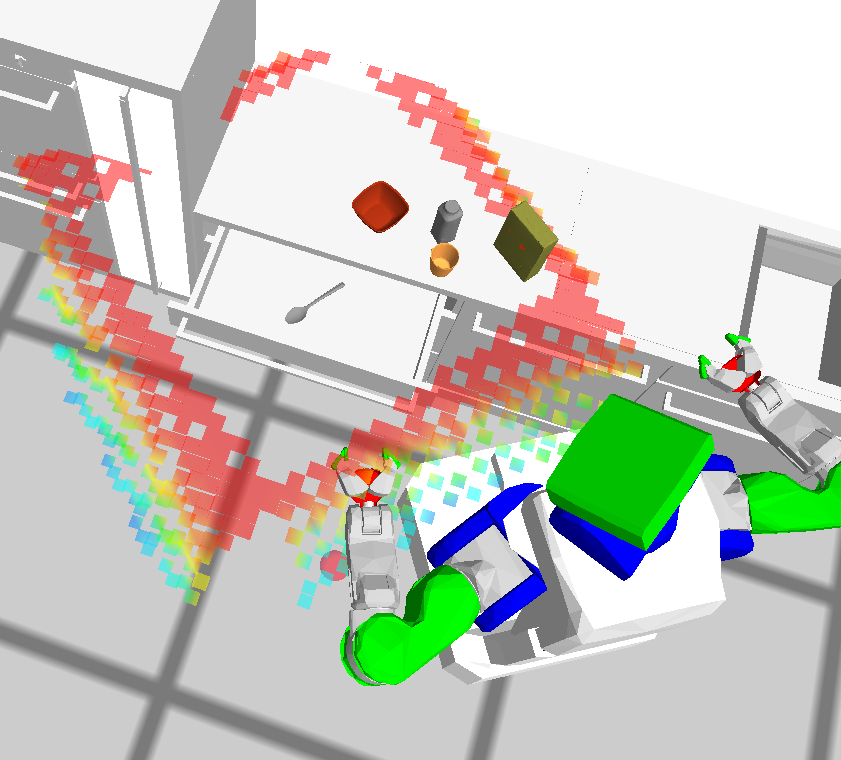}
	\includegraphics[width=0.32\columnwidth,height=0.32\columnwidth]{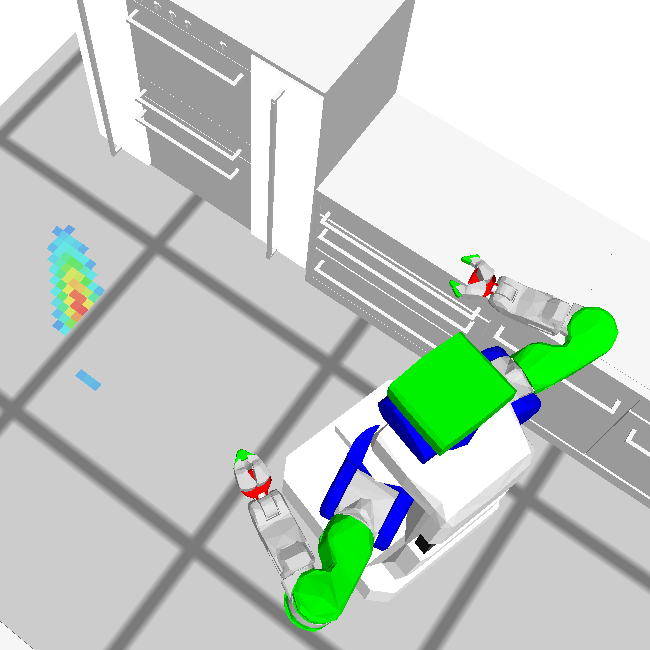}
	\caption{Robot base pose distributions, visualized as heat maps, generated by alternative approaches: (left) heuristic, (middle) general learned model, (right) specialized model.}
	\label{fig:learning}
\end{figure}

For example, to infer where the robot should stand to open the right tall vertical drawer, the Heuristics module would return  a uniform distribution, depicted in Figure \ref{fig:learning} (left) as the red highlighted region.
However, some locations from this area are more likely to succeed than the others. Additionally, as we use a general heuristic that can be applied for opening/closing any door or drawer in the environment, some suggested locations might always lead to failed execution for the particular drawer.
Therefore, finding a good location by sampling from this distribution is expensive.

As an alternative, we implemented an experience-based Specialization module for faster parameter inference.
To collect training data for the experience-based parameter inference models, we log all the information that is available to the robot during the execution of the experiments.
We call these experience logs \textit{episodic memories}\footnote{\scriptsize \url{https://ease-crc.github.io/soma/}}.
They contain not only the low-level data streams, such as joint positions or camera images, but also the semantically annotated steps of the plan and their relations.
As the data is so comprehensive, it can be used not only for learning specific action parameters but for solving any learning problem, even a future one, defined long after the data acquisition.
Please note that learning sets of parameters for single actions, as opposed to an end-to-end policy, makes our system more transparent and more scalable to new contexts.
To get as much data as possible in a short period of time, we use the fast plan projection simulator, based on the Bullet physics engine \cite{projection}. 

We have two types of models that we train with the acquired data: the general models and the specialized ones.
The general models, which we introduced in our previous work \cite{ralf}, abstract away from the environment and can, therefore, deal with the dynamically changing environments.
For example, Figure \ref{fig:learning} (middle) shows a distribution, generated by the general model, for robot base poses to grasp a spoon located anywhere in the environment with the learned best grasp for this object.
Please note that the learned distribution is not uniform anymore.
We showed in \cite{ralf} that with this model the number of execution failures on the real robot is reduced by 63\%.
As you can see, due to its generality, this distribution contains positions that would result in collision of the robot with the environment.
Therefore, we combine the learned distributions with the ones based on heuristics, by applying multiplication and normalization.

Creating the general models requires a large amount of training data with a comprehensive set of variations therein and a lot of context knowledge, which is acquired by logging the internal world state of the robot during data collection. For the purposes of the household marathon demo, we developed simpler models, which are specialized towards the given environment and the given robot. Our assumption is that the large furniture pieces rarely change their locations in the environment and the manipulated objects are typically found at their storage locations, e.g., in a specific household milk could typically be stored in the fridge door. 
Thus, instead of learning a distribution for grasping an object located anywhere in the environment, we train a model for grasping it, e.g., from a specific drawer.
Same goes for manipulating articulated objects.
%
Figure \ref{fig:learning} (right) shows the distribution of robot base poses to open the right vertical drawer, as generated by the specialized model.
This is a multivariate Gaussian distribution trained for this specific drawer specifically for the PR2.
We created similar distributions for all the objects the robot needed to interact with in the scope of the household marathon experiment. 
Using these models, the number of execution failures reduced by 81\%.

%% file: perception.tex
The Perception system is called RoboSherlock\footnote{\scriptsize \url{http://robosherlock.org} \hspace{0.5ex} \url{https://github.com/robosherlock}} \cite{robosherlock}. It is built upon three core principles:
\begin{itemize}
\item It adopts the \textit{Unstructured Information Management Architecture} \cite{watson} to formulate the problem of perception as information extraction from unstructured data.
\item Perception is \textit{taskable} in Robosherlock, i.e.\ depending on the object description received from the Plan Executive, RoboSherlock builds a custom knowledge-based perception pipeline for each received query \cite{taskable}.
\item Since there is no single algorithm that can solve all perception tasks, RoboSherlock uses an \textit{ensemble of experts} approach, where each expert solves a specific problem, e.g., color segmentation, 3D registration, transparent object perception, etc.
\end{itemize}

One of the crucial experts is RobotVQA \cite{kenghagho20paperiros},
which is a scene-graph and Deep-Learning-based visual question answering system for robot manipulation. At the heart of RobotVQA lies a multi-task Deep-Learning model that extends Mask-RCNN, and infers formal semantic scene graphs from RGB(D) images of the scene at a rate of $\approx5$ \textit{fps} with a space requirement of $5.5GB$. 
%
%
%
The graph is made up of the set of scene objects, their description (category, color, shape, 6D-pose, material, 
mask) and their spatial relations. Moreover, each of the facts in the graph is assigned a probability as a measure of uncertainty. 
To represent the scene graphs, the probabilistic and decidable description logic with concrete domains and loop relations P-ALCNHr+(D) is used.

\begin{figure}[htb]
	\centering
	\includegraphics[trim=0 165 0 155,clip,width=\columnwidth]{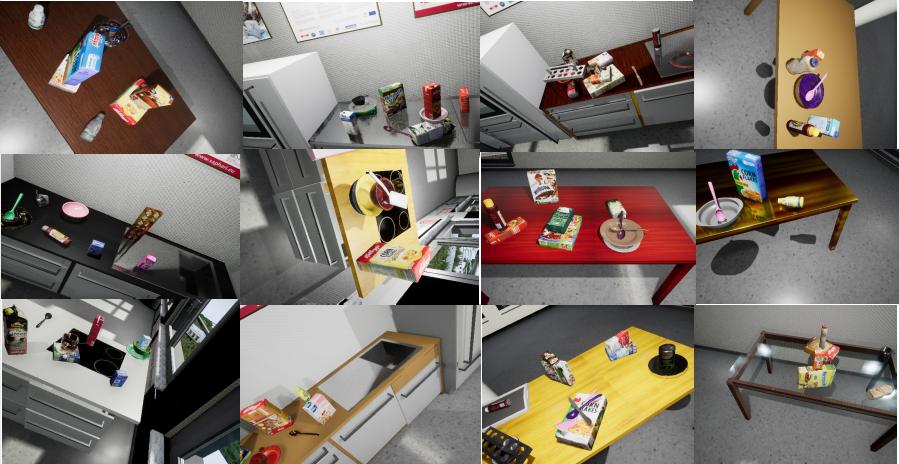}
	\caption{Situated and embodied simulated scenes}
	\label{fig:data}
\end{figure}

To train our Deep Learning model, we implemented a generator of situated and embodied scenes 
within Unreal Engine 4\footnote{\scriptsize \url{https://www.unrealengine.com}}, a photo-realistic and physics-faithful game engine (see Figure \ref{fig:data}). 
We add background noise to the scenes by augmenting the virtual data with 
objects that are not a part of the manipulation task 
such as cables, chairs and pictures.

\begin{figure}[htb]
	\centering
	\includegraphics[width=\columnwidth]{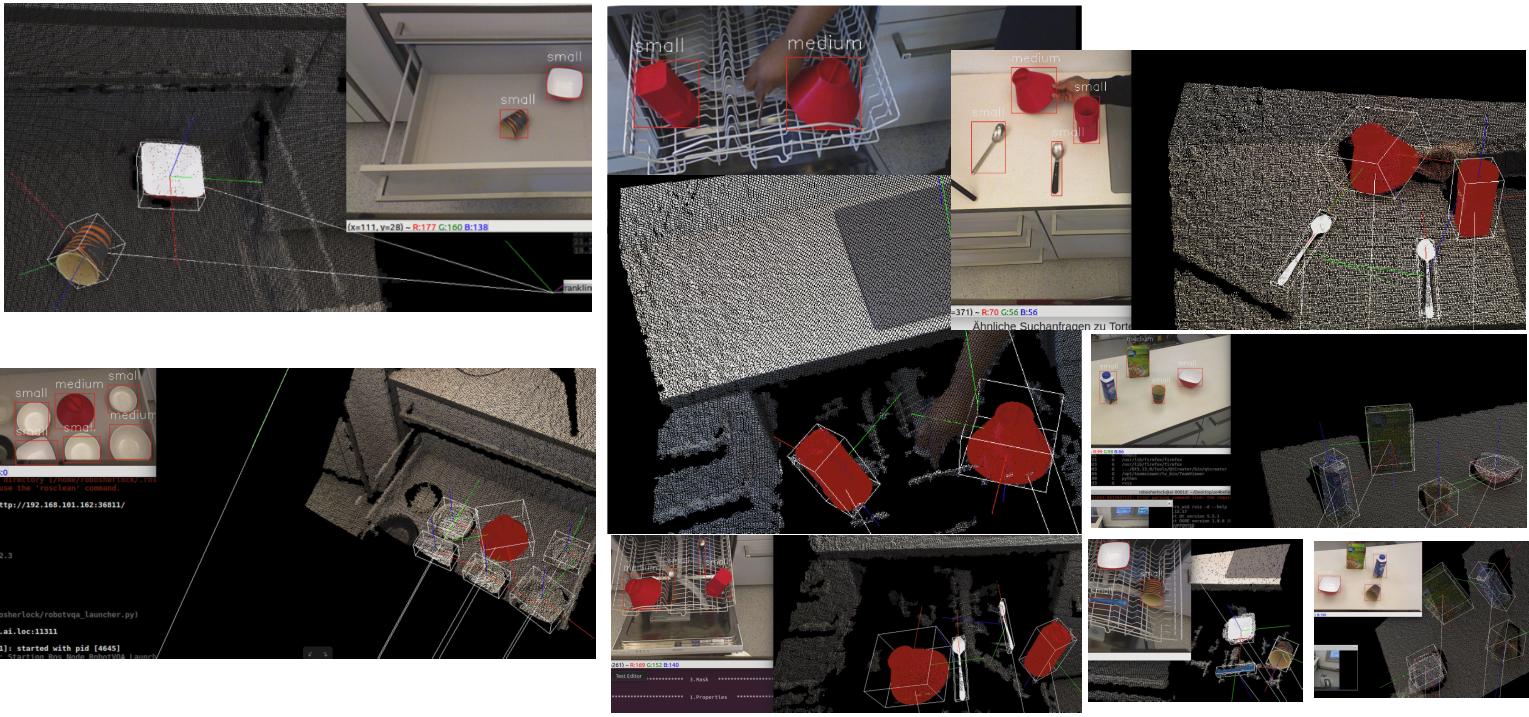}
	\caption{Pose estimation of arbitrarily oriented objects.}
	\label{fig:pose2}
\end{figure}

During the training of the multi-task Deep-Learning model with multi-loss, the learning progress on the pose estimation was very slow due to object symmetries and computational conflicts (spatial relations are rotation-invariant, while the 6D-pose is not). This was not sufficient for the purposes of the household marathon experiment. Thus, we added an additional step to the perception pipeline: 
after computing the visual mask of the object, we extract the point cloud of the object from the depth map and use PCA to compute its three main coordinate axes. To correctly label the axes, 
we start 12 parallel instances of ICP, one for each of the 12 possible axis labeling choices. Then we select the ICP result with the best score. The results are illustrated in Figure \ref{fig:pose2}.

%% file: motion_planning.tex
The Motion Planning framework we use is called Giskard\footnote{\scriptsize \url{http://giskard.de} \hspace{1em} \url{https://github.com/SemRoCo/giskardpy}}.
It allows the user to define multiple simultaneous goals and generates whole body motions, which means that we plan for PR2's arms, torso, base and head simultaneously.

To perform an action on the robot, the Plan Executive sends a motion description to the Motion Planner, e.g.,
the motion to place an object onto a table looks as following:

\begin{lstlisting}[style=Lisp]
(a motion
   (type moving-arm)
   (goal-poses (left-tool-frame ~some-pose-stamped~))
   (collision-mode allow-fingers-and-object)
   (collision-object kitchen-furniture)
   (collision-object-part dining-table)
   (constraints (keep-vertical-orientation look-at-hand)))
\end{lstlisting}

The \textit{goal-poses} can contain poses for different end effectors, e.g., for a dual-arm motion.
The \textit{collision-mode} can be any of \textit{avoid-all}, \textit{allow-all}, \textit{allow-arm}, \textit{allow-hand}, \textit{allow-fingers} or \textit{allow-fingers-and-object}, which specifies if collisions are allowed between the robot body parts or the grasped object and whatever is specified in the \textit{collision-object} parameter. Additional \textit{constraints} can also be specified, e.g., to keep the grasped object parallel to the floor, the gaze fixed on the hand, the joints pulled towards the center of their limit range, etc.
Before sending the motion description to the Motion Planner, the Plan Executive uses a physics engine to ensure that the goal pose is reachable from the current robot pose and does not cause any unwanted collisions \cite{projection}.
The motion description is then translated by Giskard into a list of mathematically defined goals. 

Goals are defined as a task function $f(o)$ with limits on its first derivative.
The vector of observable variables, $o$, contains the robot joint positions, but can also contain goal parameters, poses of environment objects, contact information, etc.
This gets turned into a quadratic programming constraint optimization problem to compute joints velocities that approach the desired velocity limits for the task function as closely as possible.
This process is explained in detail in our previous work \cite{fang2016learning}.


Giskard offers an API for defining these task functions to represent goals.
It can parse a URDF\footnote{\scriptsize \url{http://wiki.ros.org/urdf}} description of the robot to create forward kinematic expressions for any kinematic chain of the robot.
In addition, it manages $o$ and offers symbols that can be used in the task function definition, for anything that is a part of the internal world state. 
The environment therein is also represented in the URDF format.

For the purposes of the household marathon experiment, we have implemented a number of general goals,
including the ones for opening/closing doors and drawers as well as goals for each of the possible constraints that can be used in the \textit{constraints} parameter mentioned above.

The collision avoidance goals are added automatically in Giskard. They can be parameterized by the above-mentioned \textit{collision-mode} variable.
The Bullet physics engine \cite{bullet} is employed to compute closest points between the robot links or the robot and the environment.
These points and the contact normal are then used to apply a force that pushes the link away from the contact.

To prioritize between multiple goals, a weight can be assigned to each goal, specifying its precedence in case of conflicts with other goals.
The weight management is centered around the collision avoidance, because, in practice, the most common conflict is between the goals for interacting with the environment and avoiding collisions.
Therefore, the collision avoidance is implemented with two thresholds:
one, which cannot be exceeded (its default value is $0m$) and the other one keeps a collision distance of, e.g., $0.05m$ with a fixed goal weight $w_{c}$.
There are two more defined weights: $w_{ca}$ is bigger than $w_{c}$, used for goals that interact with the environment, and $w_{cb}$ is below $w_{c}$, used for any other goal.